%% file: CCR.tex
\documentclass[twoside,twocolumn]{article}

\usepackage{fitee}

\usepackage[colorlinks, breaklinks = true]{hyperref}		
\addto{\captionsenglish}{%
  
}

\usepackage{amsmath,amsfonts,amsthm}
\usepackage{algorithm,algorithmic}
\usepackage[algo2e,ruled,vlined]{algorithm2e}

\input{TexFiles/math_notation}
\begin{document}

\title{On Interpretable Approaches to Cluster, Classify and Represent Multi-Subspace Data via Minimum Lossy Coding Length based on Rate-Distortion Theory}

\author[$\ddagger$1]{Kai-liang LU}%
\author[2]{Avraham Chapman}
\affil[1]{College of Automation Engineering, Guangzhou Vocational University of Science and Technology, Guangzhou 510550, China}
\affil[2]{School of Computer Science, The University of Adelaide, Adelaide 5005, Australia}

\shortauthor{Lu and Chapman}	

\authmark{}



\corremailA{lukailiang@163.com}
\corremailB{avraham.chapman@adelaide.edu.au}
\emailmark{}	

\dateinfo{Received mmm.\ dd, 2022;	Revision accepted mmm.\ dd, 2022;    Crosschecked mmm.\ dd, 2022}

\abstract{To cluster, classify and represent are three fundamental objectives of learning from high-dimensional data with intrinsic structure. To this end, this paper introduces three interpretable approaches, i.e., segmentation (clustering) via the Minimum Lossy Coding Length criterion, classification via the Minimum Incremental Coding Length criterion and representation via the Maximal Coding Rate Reduction criterion. These are derived based on the lossy data coding and compression framework from the principle of rate distortion in information theory. These algorithms are particularly suitable for dealing with finite-sample data (allowed to be sparse or almost degenerate) of mixed Gaussian distributions or subspaces. The theoretical value and attractive features of these methods are summarized by comparison with other learning methods or evaluation criteria. This summary note aims to provide a theoretical guide to researchers (also engineers) interested in understanding “white-box” machine (deep) learning methods.}


\doi{10.1631/FITEE.1000000}	
\code{A}
\clc{TP}


\publishyear{2022}
\vol{19}
\issue{1}
\pagestart{1}
\pageend{5}

\support{Project supported by the National Natural Science Foundation of China (No.~51405289)}

\orcid{Kai-liang LU, http://orcid.org/0000-0002-9293-5801}	
\articleType{}

\maketitle

\section{Introduction} \label{sec:intro}

Complicated multi-modal, multivariate distributions are ubiquitous in engineering, appearing in fields as diverse as bio-informatics, computer vision, image/signal processing and pattern recognition. This data often has an intrinsic structure that spans multiple subspaces or includes multiple substructures.

It is the goal of this essay to summarize the latest progress on dealing with such data through the use of a family of techniques revolving around the Minimum Coding Length of the data. This essay will show how to apply the Minimum Coding Length to achieve three common objectives:

\begin{itemize}
	\item \textbf{Interpolation}: Identify which samples belong to the same subspace, i.e., segmentation or clustering.
	\item \textbf{Extrapolation}: Determine to which subspace a new sample belongs, i.e., classification.
	\item \textbf{Representation}: Find the most compact and discriminative representations for sample data.
\end{itemize}

\subsection{Lossy Coding and Coding Length}

A coding scheme is designed to map data into a series of binary bits, with the idea that the original data can later be reconstructed from those binary bits. For discrete data types, such as integers, this can be done exactly. However, for non-discrete data it is necessary to establish a cut-off to the degree of encoding precision.

Take a set of vectors $V = (v_1, v_2, v_3, \dots, v_m) \in \R^{n \times m}$. A lossy encoding scheme can be defined to map $V$ to a series of bits with a maximum reconstruction error of $E[\|v_i-\hat{v}_i\|] \leq \epsilon^2$. The length of those encoded bits could then be described with the function $L(V) : \mathbb{R}^{n \times m} \rightarrow \mathbb{Z}_+$.

This concept of encoding length is the cornerstone of the many techniques discussed in this essay and will be expanded upon in Section~\ref{sec:lossycoding}.

\subsection{Segmentation as a Clustering Problem}
\label{SegmentationClusteringIntro}

Segmentation, or similarly clustering, is an important step in analyzing, compressing, interpreting and representing multi-subspace data. In simple terms, it refers to the process of sorting data together into useful groupings based on some measure of similarity. 
However, when creating a meaningful segmentation algorithm one must answer the question of how to define segmentation mathematically? One must define a useful criterion for the segmentation process. One must also have a way of measuring the quality of the segmentation -- the “gain” or “loss”. 

The literature proposes a wide variety of approaches and solutions to the questions above from the statistical domain~\cite{A1-2} and the traditional machine learning space~\cite{PRML}.

\subsubsection{Defining segmentation as a model estimation problem}
\label{sec:DefiningSegmentation}

One traditional method of defining segmentation is to cast the problem as a model estimation problem. To do so, one must select a class of models that will hopefully fit each substructure or subspace -- the simpler, the better.
Probabilistic distribution, such as the Gaussian distribution, or geometric or algebraic sets, such as linear subspaces, are common choices as models. 
If one thinks of the mixed data being segmented as being drawn from a mixture of these modelling distributions or sets, one can then attempt to estimate the necessary mixture of the models and assign each datum to its most likely model.

Viewed as a model estimation problem, segmenting the data and estimating the models are strongly coupled tasks. Many approaches to decoupling the two tasks have been proposed. They can broadly be split into two camps:

\begin{itemize}
\item Iteratively address data segmentation and model estimation. Examples of this type of method include the {\em K-means} algorithm~\cite{lsqpcm, Forgy1965ClusterAO, Jancey1966MultidimensionalGA, Macqueen67somemethods} and its variants~\cite{DeterministicAnnealing, DistanceMetricLearning, ClusteringAppearences} as well as the {\em Expectation-Maximization} (EM) algorithm \cite{ExpectionMax}. These are basically greedy descent algorithms that attempt to find the {\em maximum likelihood} (ML) estimate of a mixture of probabilistic distributions.
\item First estimate a mixture model that has no dependency on the segmentation of the sample data. Next, decompose the model into individual components. An example of this approach would be \emph{Generalized Principal Component Analysis} (GPCA) \cite{vidal2005generalized,GPCA}, which assumes the mixture model to be an arrangement of subspaces.
\end{itemize}

What all of these methods have in common is that they assume a good estimate of the data's underlying mixture model. This is critical to their success at segmenting the data. 
For example, one could assume that sample data $W = (w_1,w_2,\dots,w_m)$ was drawn from a combination of probability distributions $p(x|\theta,\pi)\doteq\sum_{j=1}^{k}\pi_jp_j(x|\theta_j)$.
\footnote{For clarity, the exact same notations as the original papers \cite{A1,A2,A3} are used.} 
In order to discover the optimal estimate for the mixture model, one must choose from any of the model estimation criteria. An example would be the \emph{maximum likelihood} (ML) {\em estimate}:
\begin{align}
	\label{eq:ml-crit}
	(\hat\theta,\hat\pi)_{ML} = 
	\arg\max_{\theta,\pi} \sum_{i=1}^{m}\log p(w_i|\theta,\pi).
\end{align}
Here, $\theta$ refers to the parameters of distribution of interest. The likelihood function of this sort of mixture model can be optimized through the use of the EM algorithm~\cite{EMAlg} or one of its variants. Criterion~\eqref{eq:ml-crit} can be seen as minimizing the negated log-likelihood $\sum_i -\log p(w_i|\theta,\pi)$. \cite{A1-17} pointed out that this approximates the expected coding length $L(W|\theta,\hat\pi)$ -- the bit count needed to store the data if it were produced using the {\em optimal coding} scheme for the distribution $p(x|\hat\theta,\hat\pi)$.

Estimation of the mixture model is further complicated by the fact that the number of component models is not often known a priori. 
In such cases, one must estimate the number from the data. This estimation can be made all the more difficult in the face of data corruption due to label noise or other sources of outliers. 
To varying degrees, any criterion used for selecting representative models can be seen as attempting to minimize the minimum coding length needed for describing both the data and generating model. This has been described as the \emph{Minimum Description Length} (MDL) criterion in the literature \cite{A1-11,A1-19,Rissanen1978,A1-20}:
\begin{align}
	\label{eq-2}
	(\hat\theta,\hat\pi)_{MDL} &= 
	\arg\min_{\theta,\pi} L(W,\theta,\pi) \nonumber \\ 
	&= L(W|\theta,\pi) + L(\theta,\pi).
\end{align}
Here, the two parameters $\theta$ and $\pi$ are assumed to be drawn from the distribution $p(\theta,\pi)$. $L(\cdot)$, the length function, would provide the optimal Shannon-coding for each parameter. 
According to~\cite{A1-17}, this would be $-\log p(W|\theta,\pi)$ for $W$ and $-\log p(\theta,\pi)$ for $\theta$ and $\pi$. This objective function is equivalent to \emph{maximum a posteriori} (MAP) estimate, which makes the EM algorithm the best method for the job.

\subsubsection{Adapting ML and MDL to the messy reality of real-valued data}
\label{AdaptingMLandMDL}

ML and MDL as discussed above are all well and good with discrete numbers, or even with continuous random variables when the quantization error is near zero. In those cases, the data can be encoded losslessly and the ML and MDL criteria correspond to the minimum coding lengths for that data.

However, for real-valued data a lossless encoding would require an infinite encoding length. To encode in a finite length, one would have to accept a certain degree of reconstruction error, $\varepsilon>0$. It would be necessary to encode the parameters of the model(s) that could generate the data along with the model mixtures for each datum, quantized to $\varepsilon$.

To address this situation, \cite{A2-25} conducted a study of the properties of both the \emph{lossy ML} (LML) and the \emph{lossy MDL} (LMDL) criteria:
\begin{align}
	\label{eq-3}
	(\hat\theta,\hat\pi)_{LML} = 
	\arg\min_{\theta,\pi} R(\hat p(W),\theta,\pi,\varepsilon), ~~\text{and}
\end{align}
\begin{align}
	\label{eq-4}
	(\hat\theta,\hat\pi)_{LMDL} = 
	\arg\min_{\theta,\pi} R(\hat p(W),\theta,\pi,\varepsilon) + L(\theta,\pi).
\end{align}
In Eq.~\eqref{eq-3} and \eqref{eq-4}, $\hat p(W)$ refers to an empirical estimate of the probabilistic distribution that generated the data $W$.
\cite{A2-25} showed that to minimize the coding rate of the data with maximum distortion $\varepsilon$ would be mathematically the same as computing the LML or LMDL estimates. This makes it a useful estimator, exhibiting strong consistency, and a natural way to measure the quality of the segmentation of real-valued mixed data.

Although a huge number of types of data mixtures are possible, this discussion is restricted to a discussion of data that consist of a mixture of multiple \emph{Gaussian-like} groups (other types of data are discussed later).
However, a wide variety of Gaussian-like distributions are compatible, including differing anisotropic covariances and even nearly {\em degenerate} data. In all cases, the goal would be to fit the data to {\em multiple subspaces}, not necessarily all of the same dimension.

Segmentation on data can be performed by minimizing the overall coding length of each data segment, subject to the aforementioned maximum acceptable distortion $\varepsilon$. 
After an analysis of the coding length of mixed data, \cite{A1} were able to make a compelling case for a connection between data segmentation and many fundamental concepts from the lossy data compression and rate-distortion theory domains. They showed that it was possible to approximate the asymptotically optimal solution for how to compress data with a {\em deterministic} segmentation algorithm. The algorithm they posited required only a single parameter, the allowable distortion. Without any parameter estimation, the algorithm could take a given distortion and determine the corresponding number of data segments and dimensionality of each group.

This is explored further in Section \ref{sec:SegmentationviaMCL}.

\subsection{Minimum Incremental Coding Length and Classification}
\label{sec:ClassificationICLIntro}

At the forefront of every textbook on statistical learning is the problem of classification \cite{A2-32,PRML}. The problem is usually posed as an attempt to create a classifier for labelled i.i.d. data drawn from an unknown probability distribution: $(\x_i,y_i)\sim P_{X,Y}(\x,y)$, where $\x_i \in \R^n$ is the $i$th observation and $\y_i \in \{1,\dots,K\}$ is $x_i$'s associated class label. The problem is solved by constructing a classifier $g:\R^n \to \{1,\dots,K\}$ that minimizes the expected risk:
\begin{align}
	\label{eq-5}
	g^* = \argmin \mathbb E[\I_{g(X)\neq Y}].
\end{align} 

This is with respect to the underlying probabilty distribution $P_{X,Y}$. In plain English, the classifier must correctly output the associated label for every input datum. When one knows both the class priors $p_Y(y)$ and the conditional class distributions $P_{X,Y}(\x|y)$, the \emph{maximum a posterior} (MAP) value
\begin{align}
	\label{eq-6}
	\hat y(\x) = \arg\max_{\substack{y \in \{1,\dots,K\}}} 
	p_{X|Y}(\x|y)\, p_Y (y)
\end{align}  
represents the optimal classifier.

\subsubsection{Issues with learning the singular distributions from finite training samples}

In a standard classification scenario, one would set out to learn the distributions $P_{X,Y}(\x|y)$ and $p_Y(y)$ from a set of training data samples and their associated class labels. Traditional approaches work on the assumption that the distributions are non-singular and that are sufficient samples. However, in many challenging classification problems in computer vision, this assumption does not hold up~\cite{GenerativeAndFisher, GradientLearningDocumentRecognition, VariableLighting, facialrecognitionframework}. 
An example would be from the realm of facial recognition, whereby there are training sets of images of a single person's face which are taken from many angles and under many lighting conditions. This data would often reside in a low-dimensional manifold of the overall embedding space \cite{ClusteringAppearences}. In such cases, the associated distributions are \emph{singular} or \emph{nearly singular}. An additional problem lies in the high dimensionality of computer vision problems' input data, which tends to make the set of training images {\em sparse}.

\cite{A2-32} showed that trying to take a sparse set of samples and infer a generating probability distribution is an inherently ill-conditioned problem. In regard to singular distributions, he further showed that Eq.~\eqref{eq-6} does not have a well-defined maximum. 
This means that it is important to {\em regularize} the distribution or its likelihood function in order to be able to use it to classify new observations or even to infer it from training data. Regularization can be done in an explicit fashion through the use of smoothness constraints. It's also possible to use parametric assumptions about the distribution for a kind of implicit regularization. Even when the underlying assumption is assumed to be Gaussian, it remains necessary to have regularization when there are not many samples available to learn from~\cite{GPCA}. 
This is a huge issue in computer vision and bio-informatics, since these areas typically deal with very high-dimensional data spaces. When the number of samples available is on the same order as the dimensionality of the data, naive covariance estimators can produce bad results~\cite{Bickel_2008} -- higher dimensionality requires a correspondingly higher number of training samples for them to work. This problem is also present with estimators of subspace structure, including principal component analysis~\cite{Zou04sparseprincipal}.

In the world of computer vision, different classes of data often have very different intrinsic complexities, resulting in them lying in subspaces or manifolds of differing dimensionality. 
For example, with facial detection, the features corresponding to the face often form a low-dimensional structure “embedded” as a sub-manifold amongst many other random features that comprise the rest of the image.

Model selection criteria such as MDL (described in Section~\ref{sec:DefiningSegmentation}) are major improvements over MAP for exactly this reason. They can estimate a model when there are many classes of widely differing complexity. 
Recall that MDL works by attempting to select a model that minimizes the overall coding length for a given set of training data. What MDL does not do, however, is to specify the best way to account for the model complexity when considering new test data and models with differing dimensions. 
Unlike model estimation as discussed above, which is focused on estimating a model from a set of training data, this is instead focused on assigning a new test sample to an already existing model.

\subsubsection{Minimum coding length and MAP's encoding interpretation}

After the process of estimating the distributions of $p_{X|Y}$ and $p_Y$ is complete, it is then possible to derive a classifier. Consider the estimates of the distributions to be $\hat p_{X|Y}$ and $\hat p_Y$. It possible to substitute those into the MAP classifier~\eqref{eq-6}. The MAP classifier can thus be expressed as
\begin{align}
	\label{eq-7}
	\hat y(\x) = \arg\max_{\substack{y \in \{1,\dots,K\}}} 
	- \log p_{X|Y}(\x|y) - \log p_Y (y).
\end{align}  
Where, $- \log p_{X|Y}(\x|y)$ calculates the number of bits needed to code sample $\x$ with respect to the distribution of class $y$. $- \log p_Y (y)$ calculates the number of bits needed to code the label $y$ associated with sample $\x$. 
Expressing the MAP classifier in this manner allows us to use an encoding interpretation for the process. The optimal classifier would ideally minimize Shannon’s optimal (lossless) coding length for the test data $\x$ with respect to the distribution of $x$'s true class. In this essay, \emph{Minimum Description Length} (MDL) criterion is followed for classification.

However, as has been described above, learning the (potentially singular) distributions $p_{X|Y}$ and $p_Y$ from just a few samples in a high-dimensional space can be very difficult. It therefore behooves us to seek out alternative methods for implementing the above criterion.

\subsubsection{The minimum incremental coding length (MICL) criterion for classification}\label{MICLIntro}

\cite{A2} proposed a method for implementing the MDL criterion descried above. They set out to calculate how efficiently a new observation could be encoded by each class in the training data. They proposed to find the class that encoded the new data in the fewest possible bits and to assign the new data a class label matching that class. They dubbed this the \emph{“minimum incremental coding length”} (MICL) criterion for classification. MICL serves as a counterpart to the MDL method for model estimation and makes a good surrogate for the MDL classification.

This MICL criterion smoothly addresses the aforementioned issues of model complexity and regularization. The fact that the coding is lossy (the test data $\x$ is encoded up to an allowable reconstruction distortion) serves as a form of natural regularization. This stands in contrast with Shannon’s optimal coding length, which pre-requires precise data on the true distributions. Instead, MICL is more like {\em lossy} MDL~\cite{A2-25}.

Section \ref{sec:MICL} describes MICL in more detail. It describes how MICL measures the difference between the total volume of the training data both with and without a new observation.

\subsection{To Learn Diverse and Discriminative Representations}

Learning classification or clustering as described above often leads to the problem of overfitting. That is, trends and structures in the training data are selected and relied upon, even when those trends do not exist in the more general population. Indeed, when attempting to solve a classification problem, only the features that are useful for that classification problem tend to get selected for inclusion in representations.

Take a random vector $\x \in \Re^D$, being drawn from one or more of $k$ distributions $\mathcal{D} = \{\mathcal{D}_j\}_{j=1}^k$. It is necessary to be able to learn the distribution from a set of i.i.d. samples $\X = [\x_1, \x_2, \ldots, \x_m] \in \Re^{D\times m}$. To avoid the overfitting problem, one would want to find a good general-purpose representation via a continuous mapping, $f(\x, \theta): \Re^D \rightarrow \Re^d$, that includes all intrinsic structures of $\x$. If the representation is inclusive enough, it can support a wide variety of downstream tasks, such as classification or clustering. 

\subsubsection{Supervised learning} 

In the case of supervised learning, the task of learning a discriminative representation is often posited as finding a mapping $f(\x, \theta):\x \mapsto \y$, parameterized by $\theta \in \Theta$, for all elements of a training set $\{(\x_i, \y_i)\}_{i=1}^m$. This has been shown to be feasibly modelled by a deep neural network in areas as diverse as vision, audio and natural language processing \cite{goodfellow2016deep}. The true label $\y_i \in \Re^k$ is represented as a one-hot vector of dimension $k$. The model is trained through backpropagation to train the network parameters $\theta$ to try and eliminate the difference between the model's output and the true label:
\begin{align}
	\min_{\theta \in \Theta} \; \mbox{CE}(\theta, \x, \y) &\doteq - \mathbb{E}[\langle \y, \log[f(\x, \theta)] \rangle] \nonumber \\	
	&\approx - \frac{1}{m}\sum_{i=1}^m \langle \y_i, \log[f(\x_i, \theta)] \rangle.
	\label{eqn:cross-entropy}
\end{align}

This has the downside that it exclusively aims to predict the labels $y$, which can be a problem if the labels are incorrect. \cite{zhang2017understanding} showed that a sufficiently large model can memorize even randomly selected labels, regardless of any latent structures in the input data. This leads to the geometric and statistical properties of the data being at best obscured and at worst removed from the feature representation vectors. 
As a consequence, the feature representation vectors would be useless for any other downstream tasks, lacking semantic meaning and failing to generalize to other domains and applications. The end goal would be to reformulate the objective above to one that explicitly learns meaningful representations for the data $x$.

\noindent \textbf{Minimal discriminative features via information bottleneck:} It is popular in recent literature to divide networks into feature extractors and downstream modules. In this conception, the feature extractor is trained to select a vector of representative latent features $\z = f(\x, \theta) \in \Re^d$ that describe the input data well enough to be discriminative amongst multiple classes. These feature vectors $z$ are then used as inputs to train downstream classification tasks, $g(z)$, for predicting the class label $y$.
\begin{equation*}
    \x   \xrightarrow{\hspace{2mm} f(\x, \theta)\hspace{2mm}} \z(\theta)  \xrightarrow{\hspace{2mm} g(\z) \hspace{2mm}} \y.
\label{eqn:discriminative}
\end{equation*}

According to the {\em information bottleneck} (IB) formulation of~\cite{Tishby-ITW2015}, the network $f(x, \theta)$ will learn $z$ to the minimum level of statistics needed to support the prediction of $y$. Put formally, the model tries to maximize the mutual information $I(z, y)$\footnote{Mutual information is defined to be $I(\z, \y) \doteq H(\bm z) - H(\bm z \mid \y)$ where $H(\z)$ is the entropy of $\z$ \cite{A1-17}.} while minimizing $I(x, z)$.
\begin{equation}
    \max_{\theta\in \Theta}\; \mbox{IB}(\x, \y, \z(\theta)) \doteq I(\z(\theta), \y) - \beta I(\x, \z(\theta)), \quad \beta >0. 
\label{eqn:information-bottleneck}
\end{equation}
Because this formulation is task dependent (e.g. existing to support the prediction of label $y$), the information extracted will be the minimum necessary toward that goal. This leads to an inability to generalize, a lack of robustness to incorrect labelling and an inability to transfer knowledge to new tasks.

\noindent \textbf{Contractive learning of generative representations:} Unlike the ground-truth requiring supervised technique above, some other techniques do not require access to labels. One family of unsupervised techniques is called {\em  auto-encoding}. This method learns a good latent representation $\z \in \Re^d$ that contains enough data to recreate the original input $x$ to within a certain tolerance. This is often done through training a decoder $f(\x, \theta)$ and a generator $g(\z, \eta)$\footnote{Auto-encoding can thus be viewed as a nonlinear extension to classical PCA \cite{Jolliffe2002}.}.
\begin{equation}
     \x \xrightarrow{\hspace{2mm} f(\x, \theta)\hspace{2mm}} \z(\theta)  \xrightarrow{\hspace{2mm} g(\z,\eta) \hspace{2mm}} \widehat{\x}(\theta, \eta).
     \label{eqn:generative}
\end{equation}
Here, $z(\theta)$ is usually obtained by training $f(\x, \theta)$ and $g(\z, \eta)$ in an end-to-end fashion. A certain definition of “compactness” is specified (geometrical, statistical, etc.) and imposed on the representations obtained. These can include such measures as dimension, energy and volume. An example would be the {\em contractive} auto-encoder  \cite{contractive-ICML11}, which penalizes local volume expansion of the learned feature vectors. This is approximated by the Jacobian $\|\frac{\partial \z}{\partial \theta}\|$.

In addition to requiring some form of compactness for the latent representation, it is also important to choose a metric for determining the degree of {\em similarity} between the decoded $\widehat{\x}$ and the original input $x$\footnote{In tasks like denoising, the $\ell^p$-norm is often used, that is $\min_{\theta,\eta} \mathbb{E}[\|\x -\widehat{ \x}\|_p]$, typically with $p = 1$ or 2.}. Alternatively, one can measure the {\em similarity} between two distributions $\mathcal{D}_{\x}$ and $\mathcal{D}_{\widehat \x}$, for example using KL divergence $\mbox{KL}(\mathcal{D}_{\x}|| \mathcal{D}_{\widehat{\x}})$\footnote{When the distributions of $\x$ and $\widehat{\x}$ are discrete and degenerate this is very difficult. In reality, this is often instead achieved using an additional disriminative network, known as a GAN~\cite{goodfellow2014generative, Hong2019-GANReview2, Wang2021-GANReview3}.}. An appropriate metric should maintain the most important information in the reconstruction, whilst allowing unimportant elements to be lost. The can prove very tricky.

It would seem that representations learned through this sort of approach should be rich enough to be able to recreate the original data. However, if the wrong regularizing heuristics on $\z$ or similarity measures on $\x$ (or $\mathcal{D}_{\x}$) are chosen, the representations may be grossly approximated and only rich enough to handle the task they were trained with \cite{contractive-ICML11,goodfellow2014generative}. Naive heuristics or inappropriate measurements can fail to capture all internal subclass structures of complicated multi-modal data
\footnote{One consequence of this is {\em mode collapse}, whereby the variation in the outputs from a model collapse to zero; see~\cite{li2020multimodal-IJCV} and references therein.}. 
This can lead to downstream failure to discriminate between amongst them during classification and clustering.

Recognising this problem, \cite{A3} proposed \emph{Maximal Coding Rate Reduction} (MCR$^{2}$). This is a measure on $\z$ that can learn feature representations that can promote multi-class discriminative properties from mixed-structure data. This approach works in both a supervised and unsupervised setting.

\subsubsection{Learning diverse and discriminative representations}
\label{sec:How2learnDDR}

Consider a set of data $\X$ of a mixed distribution $\mathcal{D}$. $\mathcal{D}$ consists of component distributions $\mathcal{D}_j$. The greater the separability of the various $\mathcal{D}_j$, the more effectively $\X$ can be classified. It is thus very important that $\mathcal{D}_j$ be easily separable or made easily separable.

\begin{figure}[tbh]
	\centering
	\includegraphics[scale=0.19]{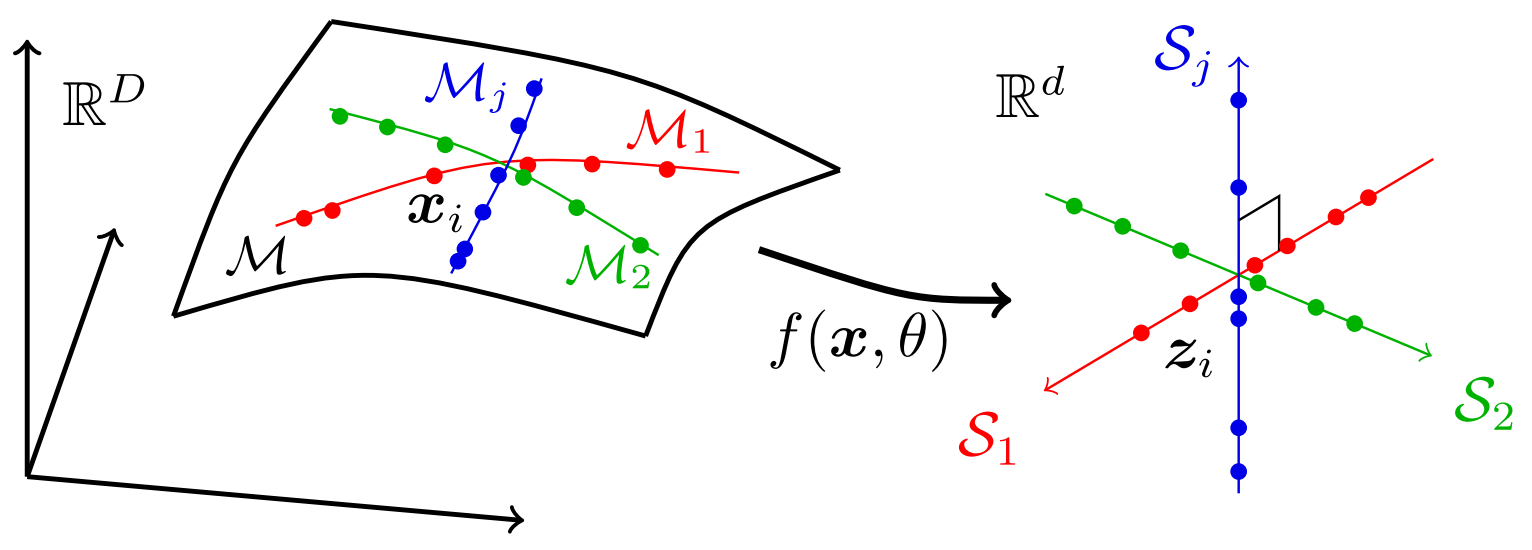}
	\caption{A distribution $\mathcal D$ with $\x\in \Re^D$, which lies on a manifold $\mathcal{M}$. Individual classes lie on lower dimensional submanifolds $\mathcal{M}_j$. MCR$^{2}$ can learn a mapping $f(\x, \theta)$ such that $\z_i = f(\x_i, \theta)$ are on a union of maximally uncorrelated subspaces $\{\mathcal{S}_j\}$ (Right)~\cite{A3}.} \label{fig:low-dim}
\end{figure}

A common assumption in the literature is that each class' inherent distribution has a relatively low-dimensional structure $\mathcal{D}_j$ \footnote{This is a realistic assumption since there is a lot of redundancy in high-dimensional data. Moreover, data from the same class should be very similar and thus highly correlated.}. This allows us to assume that each class' distribution lies on a low-dimensional submanifold $\mathcal{M}_j$, with dimension $d_j \ll D$. This would mean that the overall distribution $\mathcal D$ would lie on the union of all per-class submanifolds $\mathcal M = \cup_{j=1}^k \mathcal{M}_j$, itself embedded in a much higher dimensional ambient space $\Re^D$. This is illustrated in Figure~\ref{fig:low-dim}.

Given this assumption about the submanifolds, it would make sense to find a mapping {$\z = f(\x, \theta)$} between each submanifold $\mathcal{M}_j \subset \Re^D$ to an independent {\em linear subspace} $\mathcal{S}_j \subset \Re^d$. Figure~\ref{fig:low-dim} (middle) illustrates this. In order for that to be possible, the representations must exhibit the following properties: 

\begin{enumerate}
\item \emph{Discriminative Between-Classes}: Features vectors from different classes or clusters should exhibit very {\em low correlation} and belong to distinct low-dimensional linear subspaces.
\item \emph{Within-Class Compressibility}: Features vectors from the same class or cluster should exhibit very {\em high correlation} in the sense that they belong to the same low-dimensional linear subspace.
\item \emph{Maximum Representation Diversity}: Dimensionality (or variance) of feature vectors within each class or cluster should be \emph{as large as possible}, subject to the constraint that they stay uncorrelated from the other classes.
\end{enumerate}

Bear in mind that, while the intrinsic structures of each class or cluster are low-dimensional, they are usually not linear in their original representation $\x$. Instead, the subspaces $\{\mathcal{S}_j\}$ can be treated as \emph{nonlinear generalized principal components} for $\bm x$ \cite{GPCA}. Furthermore, equivalency between samples is task-specific. Even as samples may vary, their low-dimensional structures can still be seen as equivalent if they are {\em invariant} to domain deformations or augmentations $\cT = \{\tau \}$\footnote{So $\x \in \mathcal{M}$ iff $\tau(\x) \in \mathcal{M}$ for all $\tau \in \cT$.}. These deformations can have sophisticated geometric and topological structures that can be difficult to learn, even for convolutional neural networks (CNNs)~\cite{Cohen-ICML-2016, cohen2019general}.

This is explored further in Section \ref{sec:MICL}.

\section{Lossy Coding and Coding Length Revisited}\label{sec:lossycoding}

Consider a random vector $v$ which contains i.i.d. samples $v_i \in \R^n$ with a probabilistic distribution $p(v)$. The \emph{optimal coding scheme} and the \emph{optimal coding rate} of said random vector have an established basis in information theory~\cite{A1-17}. 
However, a more realistic and less theoretical situation would be $W = (w_1,w_2,\dots,w_m)$, which is a {\em finite} set of vectors. This dataset can be seen as a non-parametric distribution -- each vector $w_i$ in $W$ occurs with the same probability ($1/m$). This means that the optimal coding scheme described in the literature is not optimal for $W$, nor is the formula for the coding length appropriate in this setting. However, there are still lessons to be learned from information theory that can be applied to this non-parametric setting. This wisdom can be used to form a {\em tight bound} on the {\em coding length/rate} for the given data $W$, which will be described in the following section\footnote{In Appendix A of \cite{A1}, they give an alternative way to derive the upper bound. Both this and the alternative have much in common. They arrive at the same estimate, they both are good approximations for the asymptotically optimal rate-distortion function for Gaussian sources and they both utilize lossy coding of subspace-like data.}.

\subsection{Calculating Rate-Distortion}
\label{sec:Rate-Distortion}

In order to measure the quality of any proposed coding scheme, it is necessary to find ways to describe the level of compression and segmentation for that scheme. This subsection describes a Rate-Distortion function, proposed by \cite{A1} that should do the trick.

Consider a set of data $W = \{(w_i)\}_{i=1}^m$ with each $w_i \in \mathbb R^n$ and a zero mean, thus $\mu \doteq \frac{1}{m}\sum_{i}w_i=0$\footnote{Refer to Appendix B in \cite{A1} for the more complex case where the dataset's mean is not zero.}. If $\varepsilon$ is the allowable encoding and reconstruction error for $w_i$, then $\hat w_i$ is an approximation of $w_i$ with error $\mathbb E[\| w_i - \hat w_i\|] \leq \varepsilon^2$. This means that the average allowable squared error per entry $w_i$ would be $\varepsilon^2/n$.

{\em Sphere packing} can be introduced as a concept that helps to illuminate the nature of the problem of coding the vectors in $W$, subject to the mean squared error $\varepsilon^2$. Indeed, this is commonly cited in works on information theory \cite{A1-17}. Each vector $w_i \in W$ can be perturbed by up to $\varepsilon$, allowing it to exist anywhere within a sphere of radius $\varepsilon$ in $\mathbb R^n$. This error can be generally modelled as independent Gaussian noise:
\begin{align}
	\label{eq-8}
	\hat w_i = w_i + z_i, \quad \text{with} \quad
	z_i \sim \N \Big( 0,\frac{\varepsilon^2}{n}I \Big). 
\end{align}  

\begin{figure}[!tbh]
\centering
\includegraphics[scale=0.32]{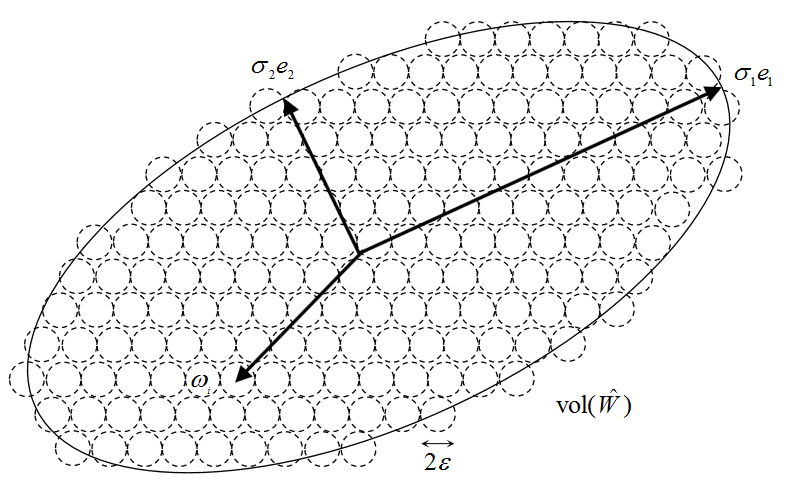}
\caption{\small Diagram of possible coding of a set of vectors that span a region $\mathbb R^n$. Their accuracy is $\varepsilon^2$, which defines the size of the spheres. To reference a vector $w_i$, one simply needs to have the label of the sphere corresponding to the vector. $e_1$ and $e_2$ in the diagram are the singular vectors for the matrix $\hat{W}$, with $\sigma_1$ and $\sigma_2$ representing the singular values~\cite{A1}.}
\label{fig:Fig1_spherepacking}
\end{figure}

This makes the covariance matrix of the vectors $\hat w_i$:
\begin{align}
	\label{eq-9}
	\hat \Sigma \doteq \Bigg [\frac{1}{m}\sum_{i=1}^{m}\hat w_i \hat w_i^T\Bigg ]
	= \frac{\varepsilon^2}{n}I + \frac{1}{m}WW^T \in \R^{n\times n}. 
\end{align} 

The volume of the region spanned by all vectors in $W$ is proportional to the square root of the determinant of this covariance matrix:
\begin{align}
	\label{eq-10}
	\text{vol}(\hat W) \propto \sqrt{\det 
		\Big( \frac{\varepsilon^2}{n}I + \frac{1}{m}WW^T \Big)}. 
\end{align}  

Similarly, the volume spanned by each random vector $z_i$ is proportional to
\begin{align}
	\label{eq-11}
	\text{vol}(z) \propto \sqrt{\det\Big(\frac{\varepsilon^2}{n}I\Big)}. 
\end{align}

If one partitions the region spanned by all vectors in $W$ into non-overlapping spheres of radius $\varepsilon$, they can then assign each of the vectors to a sphere. Assuming the use of binary numbers to label the spheres, the number of bits needed to perform the encoding can be expressed as:
\begin{align}
	\label{eq-12}
	R(W) &\doteq \log_2(\#\; \text{of spheres})
	= \log_2(\text{vol}(\hat W)/\text{vol}(z)) \nonumber \\
	&= \frac{1}{2}\log_2\det\Big(I + \frac{n}{m\varepsilon^2}WW^T\Big) 
\end{align}  

Consider what would happen if the samples in $W$ were drawn from a Gaussian distribution $\N(0, \Sigma)$. This would mean that $\frac{1}{m}WW^T$ converges to the covariance of the Gaussian distribution, $\Sigma$. It follows that $R(W)\to \frac{1}{2}\log_2\det(I + \frac{n}{\varepsilon^2}\Sigma)$ as $m\to\infty$. If $\frac{\varepsilon^2}{n}\le\lambda_{\min}(\Sigma)$, Eq.~\eqref{eq-12} becomes a very good approximation for the optimal rate distortion.

A very accurate optimal rate distortion calculation would generally require a complicated formula using a reverse water-filling algorithm on the eigenvalues of $\Sigma$ (see Theorem 13.3.3 in \cite{A1-17}). However, Eq.~\eqref{eq-12} stands as a good {\em upper bound} that holds for all $\varepsilon$. It is fairly accurate when $\varepsilon$ is small, as compared with the eigenvalues of the covariance matrix.

Another way to look at Eq.~\eqref{eq-12} would be a calculation of the rate distortion for source $W$ after being regularized with a noise of variance $\frac{\varepsilon^2}{n}$ (see Eq.~\eqref{eq-8}). The covariance ($\hat\Sigma$) of the set of thus-perturbed vectors ($\hat w_i$) would satisfy the condition $\frac{\varepsilon^2}{n}\le\lambda_{\min}(\hat\Sigma)$. This provides a handy expression of the rate distortion for all values of $\varepsilon$.

It should be noted again that Eq.~\eqref{eq-12} is only accurate as the number of samples becomes very large and the error $\varepsilon$ becomes very small. This means that Eq.~\eqref{eq-12} is not an actual coding scheme. However it is only necessary to show that such an encoding scheme can {\em in principle} attain the optimal rate $R(W)$.

\subsection{The Coding Length Function}

Recall that the coding rate was introduced in the previous section as $R(W)$. Using $R(W)$, it is possible to calculate the bit count that would be necessary to encode the $m$ vectors that comprise $W$:
\begin{align}
	\label{eq-13}
	mR(W) = \frac{m}{2}\log_2\det\Big(I + \frac{n}{m\varepsilon^2}WW^T\Big). 
\end{align}  

In addition, one would need even more bits to represent the coding method used. This can be done by specifying the singular values or vectors of $W$, which make up the principal axes of the region spanned by the data. See Figure~\ref{fig:Fig1_spherepacking} for intuition. Since there are $n$ principal axes, it would be necessary to use a further $nR(W)$ bits to encode them. Thus, encoding the $m$ vectors $W \subset \R^n$, subject to the squared error $\varepsilon^2$, would require\footnote{Consider the MDL criterion~\eqref{eq-2}. It stands to reason that if the term $mR(W)$ varies with the coding length for the data, it will also vary with to the coding length for the model parameters $\theta$.}
\begin{align}
	\label{eq-14}
	L(W)&: \R^{n\times m}\to \mathbb Z_+ \doteq (m+n)R(W) \nonumber \\
	&= \frac{m+n}{2}\log_2\det\Big(I + \frac{n}{m\varepsilon^2}WW^T\Big). 
\end{align}  

The properties of this coding length function are described in detail in Section 3.3 of~\cite{A1} or Appendix A of~\cite{A3}.

To recapitulate, a lossy coding scheme is designed to map a set of vectors $W = (w_1,w_2,\dots,w_m) \in \R^{n\times m}$ into a sequence of bits and to do it in a manner which allows for the original vectors to be recoverable, subject to a maximum allowable distortion $\mathbb E[\| w_i - \hat w_i\|] \leq \varepsilon^2$. The length of the encoded sequence is denoted as the function $L(W): \R^{n\times m}\to \mathbb Z_+$ (Eq.~\eqref{eq-14} above).

\section{Segmentation via Minimizing the Coding Length}
\label{sec:SegmentationviaMCL}

\subsection{Segmentation via Data Compression}

As described in the previous section, a set of samples $W=(w_1,w_2,\dots,w_m)\in \R^{n\times m}$ can be viewed as being drawn from a single Gaussian source. $W$ can then be encoded subject to distortion $\varepsilon^2$ using $L(W)$ bits. However, if instead the samples were drawn from a mixture of Gaussian distributions or subspaces, it would be better to separate $W$ into subsets corresponding to each generating distribution, such as $W=W_1\cup W_2\cup\dots\cup W_k$. One could then code them each separately. In this case, this would make the total number of bits
\begin{align}
	\label{eq-15}
	L^s(&W_1\cup W_2\cup\dots\cup W_k) \doteq \nonumber \\
 	&\sum_{i=1}^{k}L(W_i)+|W_i|(-\log_2|W_i|/m). 
\end{align}     

In Eq.~\eqref{eq-15}, $|W_i|$ refers to the size of each subset of $W$ associated with generating distribution $i$. The bit count necessary to losslessly encode the mapping between elements of $W$ and which subset they belong to can be expressed with $\sum_{i=1}^{k}|W_i|(-\log_2|W_i|/m)$\footnote{This operates under the assumption that the order of the samples is random, making entropy coding the best case for coding membership. On the other hand, if the samples are ordered in a way that has predictive value when considering the similarity of pairs of vectors (for example, neighbouring pixels in an image), the second term can be replaced with a tighter estimate.}.

If one uses a fixed coding scheme with coding length function $L(\cdot)$, the most optimal segmentation would have to be the one that minimizes the segmented coding length $L^s(\cdot)$ over all possible partitions of $W$. Given the nature of the rate-distortion function (Eq.~\eqref{eq-12}) when used with Gaussian data, it is possible to soften the objective function (Eq.~\eqref{eq-15}) by allowing probabilistic or fuzzy segmentation. However, this would {\em not} drive the expected overall coding length any shorter. A proof of this can be found in Section 4.2 of~\cite{A1}.

Eq.~\eqref{eq-15} takes the distortion $\varepsilon$ as an input. It is possible to modify the equation by adding a penalty term, for example $mn\log\varepsilon$, to the total coding length $L^s$ in order to give the optimal distortion $\varepsilon^*$\footnote{This particular penalty term is justified by noticing that $mn\log\varepsilon$ is (within an additive constant) the bit count necessary to code the residual $w-\hat w$ up to a (very small) distortion $\delta\ll\varepsilon$.}. The updated objective function, $\min_{\varepsilon}L^s+mn\log\varepsilon$, represents a variation on the original, except now it solely depends on the data (and not the distortion) as an input.

That said, it is very common to keep $\varepsilon$ a free parameter that can be set by the user. This allows users to create different segmentations of the data corresponding to different quantization scales.

\subsection{Minimizing the Coding Length}
\label{sec:MCL}

Ideally, one would want to find a minimization of the overall coding length across all partitions of the dataset. However, this can be an intractable combinatorial optimization problem when working with large datasets. However, \cite{A1} proposed a method for minimizing coding length via steepest descent (see Algorithm~\ref{Algo-1}).

\begin{algorithm}[htbp]
	\SetAlgoLined
	\SetKwInOut{Input}{input}
	\SetKwInOut{Output}{output}
	\Input{the data $W=(w_1,w_2,\dots,w_m)\in \R^{n\times m}$ and a distortion $\varepsilon^2>0$}
	initialize\; $\mathcal S := \{\mathcal S=\{w\}|w\in W\}$ \\
	\While{$|\mathcal S|>1$}{
		choose distince sets $\mathcal S_1,\mathcal S_2\in \mathcal S$ such that \\
		$L^s(\mathcal S_1\cup \mathcal S_2)-L^s(\mathcal S_1,\mathcal S_2)$ is minimal\\
		\eIf{$L^s(\mathcal S_1\cup \mathcal S_2)-L^s(\mathcal S_1,\mathcal S_2)\ge 0$}{
			break}{
			$\mathcal S := \{\mathcal S \setminus \{\mathcal S_1,\mathcal S_2\}\} \cup \{\mathcal S_1\cup \mathcal S_2\}$}
	}
	\Output{ $\mathcal S$}   
	\caption{\quad Pairwise Steepest Descent of Coding Length}
	\label{Algo-1}
\end{algorithm}

Taking a bottom-up approach, \cite{A1} proposed to start with every sample being treated as its own group. They would then repeatedly choose two groups, $\mathcal S_1$ and $\mathcal S_2$, such that merging them resulted in the maximum decrease in the coding length. Their algorithm would terminate when it is no longer possible to reduce the coding length by merging groups.

This algorithm could be performed in $O(m^3+m^2n^3)$ time, in the simple case where it creates and updates a table containing $L^s(\mathcal S_i\cup \mathcal S_j)$ groups, for all $i$ and $j$. In this case, $m$ is the sample count, and $n$ is the dimension of the space.

\cite{A1} showed this algorithm to be very effective when segmenting data that consist of a mixture of Gaussians or subspaces. It could tolerate a significant amount of outliers and could automatically determine the number of groups for any level of distortion.

Since this is a {\em greedy} descent method, it is not guaranteed to always find the global optimal solution to the segmentation problem for any given $(W,\varepsilon)$. \cite{A1-16} suggest that it may be possible to improve the algorithm's convergence by using more complicated split-and-merge strategies. It is also possible to derive the globally (asymptotically) optimal segmentation using concave optimization. However, this would have the downside that the computation time would increase exponentially with data size. At any rate, \cite{A1} found that the most important factor affecting the algorithm's global convergence is the relation between the sample density size and the distortion $\varepsilon^2$.

Readers may have noticed that the greedy merging process described in Algorithm~\ref{Algo-1} has some similarities to Ward's method and other classical agglomerative clustering methods. A big difference is that Ward's method assumes the use of isotropic Gaussians. On the other hand, Algorithm~\ref{Algo-1} can segment Gaussians with arbitrary covariances. This includes nearly degenerate distributions. \cite{A1} showed that these classical agglomerative approaches are inappropriate in that setting. This means that the change in coding length can actually be treated as a means of measuring the similarity amongst arbitrary Gaussians.

\subsection{Summary}

This section has endeavoured to describe why minimization of the coding length is an effective technique for segmenting multi-substructured mixed data drawn from from a mixture of (potentially almost degenerate) Gaussian distributions.

In many ways similar to LML and LMDL (see Section~\ref{AdaptingMLandMDL}), which also attempt to find the optimal segmentation of mixed data, this approach offers significant theoretical improvements:

\begin{itemize}
\item The other estimates discussed (ML, MDL, LML and LMDL) are {\em asymptotically} optimal -- they apply best to an infinite sequence of i.i.d. samples. This of course is not very helpful in practice, where one tends to have fewer than infinite samples. In fact, usually there are not even enough to reliably estimate the covariance. On the other hand, this coding length minimization technique can closely approximate the most optimal rate-distortion function, given a Gaussian source~\cite{A1-17} and a \emph{tight upper bound}. This remains true even when applied to realistic data samples.
\item This method of deterministic segmentation is \emph{approximately asymptotically optimal}. This makes it a very practical alternative in real life and can be used secure in the knowledge that it will roughly match the most optimal theoretical solution.
\item This method provides an explicit formula for the coding length/rate function, which makes it possible to evaluate the quality of segmentation\footnote{This applies to Gaussian sources. If one wants to compute a rate distortion function with arbitrary distributions, it is a far more difficult problem. \cite{A1-22} suggest many possible techniques in the arbitrary case.}. This method provides a practical algorithm (Algorithm~\ref{Algo-1}) that scales polynomially with both data size and dimension. Moreover, this technique effectively solves the model selection problem in the face of unknown group counts and significant outliers.
\end{itemize}

\section{Classification via MICL}\label{sec:MICL}

In Section~\ref{MICLIntro}, the concept of \emph{minimum incremental coding length} (MICL) was introduced briefly. This section aims to go into much more detail.

In Section~\ref{sec:IntroMICL} below, the general criterion of  MICL is discussed, along with a description of how it can be applied to unimodal and Gaussian distributions. Section~\ref{sec:AnalysisofMICL} will then go on to analyse the {\em asymptotic} behavior of MICL as the sample count tends toward infinity. Finally, Section~\ref{sec:KernelMICL} will discuss a kernel implementation that works for {\em arbitrary} data distributions. 

\subsection{Minimum Incremental Coding Length}
\label{sec:IntroMICL}

\subsubsection{Basic ideas}

As discussed above, if one has access to a lossy coding scheme with the coding length function $L_{\varepsilon}(\cdot)$, one could create an encoding of each sample's data class $\mathcal X_j\doteq \{\x_i:y_i=j\}$ using $L_{\varepsilon}(\mathcal X_j)$ bits. The length of the full dataset could thus be represented using the two-part function
\begin{align}
	\label{eq-16}
	\text{Length}(\mathcal X,\mathcal Y) = \sum_{j=1}^{K}
	L_{\varepsilon}(\mathcal X_j) - |\mathcal X_j|\log_2p_Y(j). 
\end{align}
Here, the first term is the length of the encoded data, while the second term counts the bits needed to optimally and losslessly encode the class labels $y_i$ for the empirical distribution of $y$.

Consider what would happen when encountering a new test sample $\x\in\R^n$ with an associated label $y(\x)=j$. If one adds $\x$ to the encoding of the existing training data $\mathcal X_j$, corresponding to the $j$th class, one could count the additional bits needed to code the pair $(\x,y)$ with:
\begin{align}
	\label{eq-17}
	\delta L_{\varepsilon}(\x,j) = L_{\varepsilon}(\mathcal X_j \cup \{\x\}) - L_{\varepsilon}(\mathcal X_j) + L(j). 
\end{align}   

The first and second terms in Eq.~\eqref{eq-17} count the number of extra bits needed to code $(\x,\mathcal X_j)$ with a maximum distortion $\varepsilon^2$. The final term, $L(j)$, counts the bits needed to losslessly code the label $y(\x)=j$. This can be seen as a “finite-sample lossy” surrogate for the \emph{Shannon coding length} in the ideal classifier~\eqref{eq-7}. With this in mind, one can derive the following classification criterion:

\begin{criterion} [Minimum Incremental Coding Length] 
	\label{Criterion-1}
	Assign new sample $\x$ to the class for which the fewest additional bits are need to code $(\x,\hat y)$ up to a maximum distortion $\varepsilon$: 
	\begin{align}
		\label{eq-18}
		\hat y(\x) \doteq \arg\min_{\substack{y=1,\dots,K}}\delta L_{\varepsilon}(\x,j).
	\end{align}
\end{criterion}

Criterion~\ref{Criterion-1} represents a general principle for classification. It is agnostic to both the choice of lossy coding scheme and the associated coding length function. However, the classification will only be effective when the shortest possible associated coding length for the given data is used.

\subsubsection{Gaussian data with non-zero mean}

Consider the coding length function $L_{\varepsilon}$, derived in Section~\ref{sec:lossycoding}. Recall that it is roughly asymptotically optimal when implicitly assuming Gaussian distributions and a coding scheme optimal for Gaussian sources. This is also requires the implicit assumption that the conditional class distributions $p_{X|Y}$ are unimodal, and can thus be well-approximated by Gaussians. \cite{A2} performed a rigorous analysis of the performance of MICL in this scenario. They demonstrated MICL's relationship with classical classifiers such as ML and MAP. They went on to show how using the same $L_{\varepsilon}$ function, MICL could be extended to \emph{arbitrary, multimodal} distributions via kernel technique implementation.

Consider a sample dataset $\mathcal X=(\x_1,\dots,\x_m)$. The mean of $\mathcal X$ would be $\hat{\bm\mu}=\frac{1}{m}\sum_{i}\x_i$. It is possible to represent the samples' deviations about that mean up to a maximum distortion $\varepsilon^2$ using $R_{\varepsilon}(\hat\Sigma)$ bits on average. The covariance could be expressed as $\hat\Sigma(\mathcal X)=\frac{1}{m}\sum_{i}(\x_i-\hat{\bm\mu})(\x_i-\hat{\bm\mu})^T$. With $m$ samples in the dataset, it is necessary to encode $m$ differences from the mean (e.g $\x_1-\hat{\bm\mu},\dots,\x_m-\hat{\bm\mu}$). This requires $mR_{\varepsilon}(\hat\Sigma)$ bits.

The optimal encoder/decoder pair as currently discussed requires a priori knowledge of the $\hat\Sigma$. If this is no longer known, it must be encoded, which adds an additional $nR_{\varepsilon}(\hat\Sigma)$ bits to the encoding length. However, the expected bit count needed to encode the mean $\hat{\bm\mu}$ of the samples can be bounded by $\frac{n}{2}\log_2(1+\frac{\hat{\bm\mu}^T\hat{\bm\mu}}{\varepsilon^2})$. \cite{A1} derived this bound based on the assumption that they can calculate the average number of bits required to encode $t\in\R$ with maximum distortion $\varepsilon^2$ using $\frac{1}{2}\log_2(1+t^2/{\varepsilon^2})$\footnote{Refer to Appendix B in~\cite{A1}.}. This represents an {\em upper bound} on the scalar Gaussian rate-distortion. This means that the bound on bit-count for the mean is at its {\em tightest} when $\hat{\bm\mu}$ is Gaussian. However, it remains valid even for general $\hat{\bm\mu}$.  

Combining the quantities above into a single formula, the bits required to encode $\mathcal X$ can be calculated with:
\begin{align}
	\label{eq-19}
	L_{\varepsilon}(\mathcal X) &\doteq
	\frac{m+n}{2}\log_2\det\Big(I+\frac{n}{\varepsilon^2}\hat\Sigma(\mathcal X)\Big) \nonumber \\
	&\quad + \frac{n}{2}\log_2(1+\frac{\hat{\bm\mu}^T\hat{\bm\mu}}{\varepsilon^2}). 
\end{align}  
This equation is divided into two terms. The foremost term counts the bits needed to encode the vectors $x_i$'s distributions about their means $\hat{\bm\mu}$. The latter term counts the bits needed to code the means themselves.

\subsubsection{Encoding the associated class label}

The label $Y$ associated with each sample is discrete and can therefore by encoded without loss. The form of the final term $L(j)$ in Eq.~\eqref{eq-17} depends on one’s prior assumptions about the distribution of the test data. If the test data's class labels $Y$ are known to have the marginal distribution $P(Y=j)=\pi_j$, then the optimal coding lengths are (within one bit):
\begin{align}
	\label{eq-20}
	L(j) = -\log_2\pi_j. 
\end{align}  

Should the test data consist of i.i.d. samples drawn from the same distribution as the training data, then it is possible to estimate $\hat\pi_j=\frac{|\mathcal X_j|}{m}$. On the other hand, if there is no prior information about the distribution of the class labels, it would be better to estimate $\pi_j\equiv\frac{1}{K}$. In this latter case, the excess coding length is solely dependent on the extra bit count needed to encode $\x$. In the same manner as the MAP classifier~\eqref{eq-6}, the choice of $\pi_j$ effectively creates an implicit \emph{prior} on the class labels.

\subsubsection{Putting it all together}

On the one hand, there exists a coding length function~\eqref{eq-19} for the samples in the dataset. On the other hand, there also exists a coding length function~\eqref{eq-20} for the labels associated with the samples in the dataset. The MICL criterion~\eqref{eq-18} can therefore be summarized into Algorithm~\ref{Algo-2}~\cite{A2} below.

\begin{algorithm}[htbp]
	\SetAlgoLined
	\SetKwInOut{Input}{input}
	\SetKwInOut{Output}{output}
	\Input{$m$ training samples partitioned into $K$ classes $\mathcal X_1,\mathcal X_2,\dots,\mathcal X_K$, and}
	\qquad\qquad a test sample $\x$ \\
	~~Prior distribution of class labels $\pi_j=|\mathcal X|/m$.\\
	Compute incremental coding length of $\x$ for each class:
	\begin{align*}
		\delta L_{\varepsilon}(\x,j) = L_{\varepsilon}(\mathcal X_j \cup \{\x\}) - L_{\varepsilon}(\mathcal X_j) - \log_2\pi_j,		
	\end{align*} 
	where
	\begin{align*}
		L_{\varepsilon}(\mathcal X) &\doteq
		\frac{m+n}{2}\log_2\det\Big(I+\frac{n}{\varepsilon^2}\hat\Sigma(\mathcal X)\Big) \nonumber \\
		&+ \frac{n}{2}\log_2(1+\frac{\hat{\bm\mu}^T\hat{\bm\mu}}{\varepsilon^2}). 
	\end{align*}
	Let $\hat y(\x)=\argmin_{j=1,\dots,K}\delta L_{\varepsilon}(\x,j)$.
	
	\Output{ $\hat y(\x)$}   
	\caption{\quad The MICL Classifier}
	\label{Algo-2}
\end{algorithm}

\noindent \textbf{Relationship to ML/MAP:}
It is possible to use a fully Bayesian approach to model estimation, estimating the posterior distributions over the model parameters. This even has some performance gains over ML and MAP in the finite sample realm. However, it falls short if the sample count is smaller than the count of free parameters in the model -- which is quite common in high-dimensional data. 
In that situation, the result depends very much on the choice of prior. On the other hand, MICL does {\em not} require that the sample count be larger than the dimensionality. This makes it a far superior approach when in the few-sample domain. In fact, the sections below will show that MICL is equivalent to the Bayesian approach in the asymptotic case. This leads to the question of what precise relationship MICL and MAP have with each other. Moreover, in exactly what circumstances is MICL the superior option?

\subsection{Asymptotic Convergence of MICL and its Advantages}
\label{sec:AnalysisofMICL}

This subsection delves more deeply into the asymptotic behaviour of MICL as the sample count, $m$,  tends toward infinity. Moreover, the situations under which MICL is the superior technique are identified.

\subsubsection{Asymptotic behavior and convergence rate}

Asymptotically, classification using the incremental coding length is equivalent to using a {\em regularized} version of MAP (or ML) with the addition of a reward on the dimensionality of the classes. The following theorem describes the precise correspondence\footnote{For detailed proof, refer to Appendix A of~\cite{A2}.}.

\begin{theorem}[Asymptotic MICL~\cite{A2}]
	Let some training samples $\{(\x_i,y_i)\}_{i=1}^m\sim p_{X,Y}(\x,y)$ be i.i.d. with $\bm\mu\doteq\mathbb E[X|Y=j], \Sigma_j\doteq Cov(X|Y=j)$\footnote{Assume that the first and second moments of the conditional distributions exist.}.
	As $m\to\infty$, the MICL criterion coincides (asymptotically, with probability approaching certainty) with the decision rule
	\begin{align}
		\label{eq-21}
		\hat y(\x) \doteq \arg\max_{\substack{y=1,\dots,K}} 
		&\mathcal L_G \Big(\x|\bm\mu_j,\Sigma_j 
		+ \frac{\varepsilon^2}{n}I\Big) \nonumber \\
		&+ \ln\pi_j + \frac{1}{2}D_{\varepsilon}(\Sigma_j),
	\end{align}
	where $\mathcal L_G(\cdot|\bm\mu,\Sigma)$ is the log-likelihood function for a $\mathcal N(\bm\mu,\Sigma)$ distribution\footnote{This criterion's form has a Gaussian log-likelihood. However, it works for any second-order $p_{X,Y}$, making no assumption of Gaussian properties. Nonetheless, it is not a good idea to directly apply MICL with coding length (Eq.~\eqref{eq-19}) to complicated multi-modal distributions, since this will often lead to poor classification performance. In Section~\ref{sec:KernelMICL}, the reader can find a discussion about how MICL could be modified for arbitrary data distributions.} and $D_{\varepsilon}(\Sigma_j)\doteq\trace\Big(\Sigma_j(\Sigma_j+\frac{\varepsilon^2}{n}I)^{-1}\Big)$ is the effective dimension of the $j$-th model, subject to a maximum distortion $\varepsilon^2$. 
	\label{them:AsymptoticMICL}   
\end{theorem}

This theorem shows that as the sample count tends to infinity, MICL generates a family of map-like classifiers, each parameterized by the distortion $\varepsilon^2$. Moreover, it should be noted that when all distributions are non-singular (e.g. their covariance matrices $\Sigma_j$ are non-singular) then $\lim_{\varepsilon\to0} \big(\Sigma_j+\frac{\varepsilon^2}{n}I\big)=\Sigma_j$, and $\lim_{\varepsilon\to0} D_{\varepsilon}(\Sigma_j)=N$, which is constant for all classes. Thus, in the case of non-singular data with $\varepsilon=0$, asymptotic decision boundaries created by MICL will include the conventional MAP classifier~\eqref{eq-6}. This means that if a rule for choosing the distortion $\varepsilon^2$ given a finite number of samples has the behaviour that $\varepsilon\to0$ as $m\to0$ and $\varepsilon(m)$ does not decrease too quickly, then the useful limiting behaviour described in Eq.~\eqref{eq-21} will dominate. Thus $\hat y(\x)$ will converge to the asymptotically optimal MAP criterion.

Theorem~\ref{them:AsymptoticMICL} is only strictly valid as $m\to\infty$. It does not make it clear whether one should expect to observe such behavior in real life. The following result \footnote{Proven in Appendix B of~\cite{A2}.} shows that the MICL discriminant functions, $\delta L_{\varepsilon}(\x,j)$ converge quickly to their limiting form, $\delta L_{\varepsilon}^{\infty}(\x,j)$:

\begin{theorem}[MICL Convergence Rate~\cite{A2}]
	As the number of samples $m\to\infty$, the MICL criterion~\eqref{eq-18} converges to its asymptotic form, Eq.~\eqref{eq-21}. The rate of convergence is $m^{-1/2}$. Assuming that the fourth moments $\mathbb E[\|\x-\bm\mu\|^4]$ of the conditional distributions exist, the probability is at least $1-\alpha$, $|\delta L_{\varepsilon}(\x,j)-\delta L_{\varepsilon}^{\infty}(\x,j)|\le c(\alpha)\cdot m^{-1/2}$ for some constant $c(\alpha)>0$.
	
	\label{them:ConvergRateofMICL}   
\end{theorem}

This theorem also shows that as the covariance becomes more singular, the constant $c$ becomes smaller. This suggests that the highest convergence speed would occur when the distributions are nearly singular.

\subsubsection{Improving on MAP}
\label{sec:map}

The sections above show that, in its asymptotic form, the MICL criterion~\eqref{eq-21} is equivalent to the MAP criterion. However, in the case where the sample count is finite or the distributions are singular, the MICL criterion behaves differently from the MAP criterion, resulting in significantly improved performance.

\noindent \textbf{Regularization and Finite-Sample Behavior:}
Consider the asymptotic MICL criterion~\eqref{eq-21}. Recall that its first two terms share a similar form with the MAP criterion, having a $\mathcal N(\bm\mu_j,\Sigma_j+\frac{\varepsilon^2}{n}I)$ distribution and prior $\pi_j$. This can be thought of as having a {\em regularizing} or {\em softening} effect on the distribution, to the extant of $\frac{\varepsilon^2}{n}$ on each dimension. This has an important flow-on effect. The associated MAP decision rule becomes well-defined, even with a true data distribution which is almost singular. In fact, \cite{A2} showed empirical evidence that even with {\em non-singular} distributions an appropriately chosen $\varepsilon$ could lead to $\hat\Sigma+\frac{\varepsilon^2}{n}I$ giving a more stable finite-sample estimate of the covariance. The happy result of this of course is a reduction in classification errors.

\noindent \textbf{Dimension Reward:} 
The asymptotic MICL criterion~\eqref{eq-21} contains an effective dimension term $D_{\varepsilon}(\Sigma_j)$. This can be rewritten as $D_{\varepsilon}(\Sigma_j)=\sum_{i=1}^{n}\lambda_i/(\frac{\varepsilon^2}{n}+\lambda_i)$, where $\lambda_i$ is the $i$th eigenvalue of $\Sigma_j$. An important feature of this is that the data distribution inhabits a perfect subspace of dimension $d$, e.g. $\lambda_1,\dots,\lambda_d \gg \frac{\varepsilon^2}{n}$ and $\lambda_{d+1},\dots,\lambda_n \ll \frac{\varepsilon^2}{n}$. This means that $D^{\perp}$ will lie very close to $d$. In fact, $D$ can be thought of as a {\em “softened”} estimate of the dimension, relative to the distortion $\varepsilon^2$. Writing from the perspective of ridge regression, \cite{A1-2} refer to this as the “effective number of parameters”. This means that the MICL criterion yields greater reward for distributions of a relatively higher dimension. However, one should note that the regularization induced by $\varepsilon$ has a strong “reward” for lower-dimensional distributions. This somewhat counteracts the high-dimensional “reward” from the asymptotic MICL criterion.

\subsection{Kernel Implementation for Arbitrarily Distribution}
\label{sec:KernelMICL}

The previous section considered the MICL criterion in the context of a Gaussian distribution or distributions. The analysis showed many useful properties in that case. However, a Gaussian distribution is often not the case in the real world. In fact, it is pretty rare to even know the general shape of the underlying distribution. If one knew beforehand, one could always carry out another analysis, similar to the one above with Gaussians, to reveal how the MICL criterion behaves in that case. But since this is also not realistic, this subsection will introduce a practical method for modifying the MICL criterion to work with {\emph arbitrary} distributions. Moreover, the modified method continues to preserve the desirable properties introduced above.

Given that $\mathcal X\mathcal X^T$ and $\mathcal X^T\mathcal X$ contain the same non-zero eigenvalues, one can derive the following identity:
\begin{align}
	\label{eq-22}
	\log_2\det \Big(I+\frac{n}{\varepsilon^2m}\mathcal X\mathcal X^T\Big) = \nonumber \\
	\log_2\det \Big(I+\frac{n}{\varepsilon^2m}\mathcal X^T\mathcal X\Big). 
\end{align} 

It is therefore possible to execute the coding length function~\eqref{eq-19} using only the inner products between the samples. If the samples $\x$ for each class are not drawn from a Gaussian distribution, but it is possible to obtain a nonlinear map $\psi: \R^n\to\mathcal H$ that can create transformed sample data $\psi(\x)$ that is roughly Gaussian, it is then possible to swap out the inner product $\x_1^T\x_2$ with a new one $k(\x_1,\x_2)\doteq \psi(\x_1)^T\psi(\x_2)$. This function $k(\x_1,\x_2)$ is symmetric positive definite and is referred to in the statistical learning literature as a “kernel function”\footnote{Refer to Mercer's theorem to discover the conditions under which $k(\cdot, \cdot)$ would be considered a kernel function.}. If one chooses an appropriate kernel function, is is possible to achieve superior classification accuracy with some classes of non-Gaussian distribution. The two most popular kernels in practice are the polynomial kernel ($k(\x_1,\x_2)=(\x_1^T\x_2+1)^d$) and the radial basic function (RBF) kernel ($k(\x_1,\x_2)=\exp(-\gamma\|\x_1-\x_2\|^2)$), as well as many variants thereof. By swapping $\x_1^T\x_2$ with $k(\x_1,\x_2)$, the classification operation on the test sample $\x$ now works by assigning $\x$ to the class that will incur the smallest possible number of additional bits needed to code $\psi(\x)$ jointly with $\psi(\x_1)\dots\psi(\x_m)$\footnote{Furthermore, Appendix C in~\cite{A2} describes “Efficient Implementation in High Dimensional Spaces”. Appendix D in the same paper considers the mean and dimensionality of the transformed data and accounts for it to ensure that the discriminant functions are well-defined, and still correspond to a proper coding length.}.

\noindent \textbf{Comparison to SVM:} 
In many ways, the transformation described above can be seen as similar to SVM, when generalized to nonlinear decision boundaries~\cite{PRML,A1-2}. If one considers SVM's reliance on “support vectors” (nearby samples used to define the shape of the classification hypersurface), the similarity can be found in the other direction as well, with SVM being considered as another \emph{lossy compression} approach to classification. However, this comparison reveals SVM's shortcomings when dealing with degenerate data on low-dimensional subspaces or submanifolds. In this case, the set of support vectors used to define the decision hypersurface must comprise all or almost all of the training data. This is often less efficient than using MICL to go directly to the low-dimensional structures for classification. Moreover, the kernelized version of MICL is a much simpler approach than SVM. While SVM constructs a linear classification hyperplane in the kernel space, MICL can often directly exploit details of the embedded data structures.

\subsection{Summary}
\label{sec:MICLSummary}

\begin{itemize}
\item This section described a new classification criterion based on lossy data compression, called the \emph{minimum incremental coding length} (MICL) {\em criterion}. It established the {\em asymptotic optimality} of MICL for Gaussian data. MICL generates a family of classifiers, which can be likened to classical techniques such as MAP and SVM. These classifiers extend usefulness of these more classic techniques into the realm of {\em sparse} or {\em singular} high-dimensional data.
\item In Section~\ref{sec:SegmentationviaMCL}, lossy coding length was introduced as a suitable objective function when clustering or segmenting data. Moreover, Algorithm~\ref{Algo-1} described a simple but surprisingly efficient greedy method for segmenting data drawn from a series of Gaussian generating models or linear subspaces.
\item \cite{A2} showed that the MICL criterion and its kernelized version performed competitively in the practical area of computer vision problems. In fact, it was nearly optimal in the face recognition area. These results were obtained without any domain-specific engineering. They further suggest that MICL's performance is due to its ability to automatically exploit \emph{low-dimensional structures} in {\em high-dimensional} (imagery) {\em data} for classification purposes.
\end{itemize}

\section{Learning Useful Representations via MCR\textsuperscript{2}}
\label{sec:MCR2}

\subsection{Measure of Compactness for a Representation}
\label{sec:lossy-coding}

Recall that Section~\ref{sec:How2learnDDR} described three desirable properties for a latent representation $\z$: \emph{Discriminative Between-Classes}, \emph{Within-Class Compressibility} and \emph{Maximally Diverse Representation}.

These may be highly desirable, but are not easy to obtain. It is not even clear that the properties are mutually compatible and can all be achieved simultaneously. Even if they are, is it possible to find a simple object that be used to measure the quality of candidate representations against these three properties? To do so, it would be necessary to find a way to measure the \emph{compactness} of a distribution of a random variable $\z$ from a {\em finite} set of samples $\Z$:

\begin{itemize}
\item The measure would need to be able to accurately characterize the intrinsic statistical and geometric properties of the distribution, with reference to its intrinsic dimension and volume.
\item The measure would need to not explicitly depend on the class labels associated with the data. This would allow it to be useful in all settings, including supervised, self-supervised, semi-supervised and unsupervised settings.
\end{itemize}

\subsubsection{Low-dimensional degenerate distributions} 

Information theory provides entropy $H(\z)$ as a candidate measure. However, entropy does not fit the bill, since it is not well-defined for continuous random variables with degenerate distributions\footnote{The same difficulty applies to the use of mutual information $I(\x, \z)$ for degenerate distributions.}.

Fortunately, \cite{A1-17} already proposed {\em rate distortion}, which was covered in more detail back in Section~\ref{sec:Rate-Distortion} about lossy data compression. This can be used to measure “compactness”. Recall that rate distortion $R(\z, \epsilon)$ is the minimum number of bits needed to encode the random variable $\z$ with an expected decoding error less than $\epsilon$. \cite{rate-distortion} have shown how rate distortion can be used to explain feature selection within deep networks. However, the computation of the rate distortion for a random variable is computationally intractable, except in the case of simple distributions such as Gaussian or discrete.

\subsubsection{Rate distortion with finite samples} 

In the practical domain where there are only a finite number of samples, it often not possible to know the distribution of $\z$. This makes evaluating the coding rate $R$ very difficult. What \emph{is} available is a finite number of samples $\X = [\x_1, \ldots, \x_m]$ with learned representations $\z_{i} = f(\x_i, \theta) \in \R^{d}, i = 1,\ldots, m$. Fortunately, \cite{A1} described a way to precisely estimate the bit count required to encode a finite number of samples drawn from a {\em subspace-like} distribution. The number of bits required to encode the learned representations $\Z = [\z_1, \dots, \z_m]$ to a maximum distortion of $\epsilon$ can be calculated with the following formula\footnote{One can obtain this formula by packing $\epsilon$-balls into the space spanned by $\Z$ or by computing the bit count necessary to quantize the SVD of $\Z$ to that degree of precision. See the work by \cite{A1} for proofs.}:
$\cL(\Z, \epsilon) \doteq \left(\frac{m + d}{2}\right)\log \det\left(\I + \frac{d}{m\epsilon^{2}}\Z\Z^{\top}\right)$. 
This shows that it is possible to measure the compactness of the learned features on the whole in terms of the average coding length for each sample, since the samples size $m$ is large. Thus, the {\em coding rate} subject to the max distortion $\epsilon$:
\begin{equation}
	R(\Z,\epsilon) \doteq \frac{1}{2}\log\det\left(\I + \frac{d}{m\epsilon^{2}}\Z\Z^{\top}\right).
	\label{eqn:coding-length-eval}
\end{equation}

\subsubsection{Rate distortion of mixed distribution data}

One common property of multi-class data features $\Z$ is that they tend to belong to multiple low-dimensional subspaces. Thus, it is often more easy to first partition the data into multiple subsets $\Z = \Z_1 \cup \cdots \cup \Z_k$, each corresponding to a low-dimensional subspace, making sure that the coding rate~\eqref{eqn:coding-length-eval} is accurate for each subset. After that, it becomes much easier to evaluate the rate distortion of the overall dataset.

Consider a set of diagonal matrices whose entries encode the membership of $m$ samples into $k$ classes $\bm{\Pi} = \{\bm{\Pi}_j \in \Re^{m \times m}\}_{j=1}^{k}$\footnote{Each diagonal entry $\bm \Pi_j(i,i)$ in $\bm \Pi_j$ corresponds to the probability of sample $i$ being part of to subset $j$. This means that $\bm{\Pi}$ lies in a simplex: ${\Omega} \doteq \{\bm{\Pi} \mid \bm{\Pi}_j \ge \mathbf{0}, \; \bm{\Pi}_1 + \cdots + \bm{\Pi}_k = \I\}.$}. \cite{A1} showed that the average bit count per sample (the coding rate) would be
\begin{align}
	&R^c(\Z,  \epsilon \mid \bm{\Pi}) \doteq \nonumber \\
	&\sum_{j=1}^{k}
	\frac{\tr(\bm{\Pi}_j)}{2m}\log\det\left(\I + \frac{d}{\tr(\bm{\Pi}_j)\epsilon^{2}}\Z\bm{\Pi}_j\Z^{\top}\right).
	\label{eqn:compress-loss-eval}
\end{align}

Take note that given $\Z$, $R^c(\Z, \epsilon \mid \bm{\Pi})$ is a {\em concave} function of $\bm{\Pi}$. \cite{fazel2003log-det} showed that this makes the function $\log\det(\cdot)$ an effective heuristic for use in {\em rank minimization} problems, providing guaranteed convergence to a local minimum. Since $\log\det(\cdot)$ characterizes the rate distortion of Gaussian or subspace-like distributions very well, it can be most effective in clustering or classification of mixed data~\cite{A1,A2,kang2015logdet}.

\subsection{Criterion of Maximal Coding Rate Reduction}
\label{sec:principle-mcr2}

When considering learned features and their utility in discriminating during classification tasks, one would look for two properties. On the one hand, features of different classes or clusters should be maximally incoherent relative to each other. Taken together, they should span the largest possible volume. Thus the coding rate for the entire set $\Z$ should be tend to be very large.

Contrariwise, the learned features from within the same class or cluster should enjoy a high degree of correlation and be mutually coherent. This means that each class or cluster should only span a space or subspace of the smallest possible volume, making the coding rate of the the subset for that class tend to be as small as possible.

This means that a good representation $\Z$ of $\X$ should satisfy both conditions. A partition $\bm{\Pi}$ of $\Z$ should have the largest possible difference between the coding rate for $\Z$ and the coding rate for all the subsets:
\begin{equation}
	\Delta R(\Z, \bm{\Pi}, \epsilon) \doteq R(\Z, \epsilon) - R^c(\Z, \epsilon \mid  \bm{\Pi}).
	\label{eqn:coding-length-reduction}
\end{equation}

Consider the case where the feature mapping $\z = f(\x,\theta)$ is implemented with a deep neural network. In that case, the relation between the learned representations and the rate reduction with respect to a given partition $\bm{\Pi}$ is shown in the following diagram:
\begin{equation}
	\X 
	\xrightarrow{\hspace{2mm} f(\x, \theta)\hspace{2mm}} \Z(\theta) \xrightarrow{\hspace{2mm} \bm{\Pi},\epsilon \hspace{2mm}} \Delta R(\Z(\theta), \bm{\Pi}, \epsilon).
	\label{eqn:flow}
\end{equation}

Note that $\Delta R$ is {\em monotonic} when taken on the scale of the features $\Z$. This means that to make the reduction amount comparable between different representations\footnote{A “different representation” of the same original sample might be obtained from a separate network or from another layer of the same network.}, it is necessary to \emph{normalize} the scale of the learned features. One way to do this is to apply the Frobenius norm of each class $\Z_j$ to scale with the number of features in $\Z_j \in \mathbb R^{d \times m_j}$: $\|\Z_j\|_F^2 = m_j$. Another very common method is to normalize each feature to the unit sphere: $\z_i \in \mathbb{S}^{d-1}$. This formulation neatly explains the need for batch normalization~\cite{ioffe2015batch} when training deep neural networks. A further method for normalizing the scale of the learned representations is to ensure that the mapping of each layer of the network is approximately \emph{isometric}~\cite{ISOnet}.

Having made the representations mutually comparable, the next goal would be to learn a set of features $\Z(\theta) = f(\X, \theta)$ and their partition $\bm \Pi$ (if not given in advance) that maximize the difference between the coding rate overall and the sum of the coding rates for each class:
\begin{align}
	\max_{\theta, \bm{\Pi}} &\;  \Delta R\big(\Z(\theta), \bm{\Pi}, \epsilon\big) = R(\Z(\theta), \epsilon) - R^c(\Z(\theta),  \epsilon \mid \bm{\Pi}), \nonumber \\ &\mbox{s.t.} \quad \|\Z_j(\theta)\|_F^2 = m_j, \, \bm{\Pi} \in {\Omega}.
	\label{eqn:maximal-rate-reduction}
\end{align}

This will be referred to hereafter as the \emph{Maximal Coding Rate Reduction} (MCR\textsuperscript{2}) criterion.  An interesting aside is that when it comes to clustering, only the \emph{sign of $\Delta R$} is needed to decide whether to partition the data. This fact is taken advantage of in the {\em greedy} Pairwise Steepest Descent Algorithm~\ref{Algo-1} in Section~\ref{sec:MCL}\footnote{When clustering a {\em finite} number of samples, it is important to use the more precise measurement for coding length that was mentioned earlier. \cite{A1} provide more details.}.

\noindent \textbf{Relationship to information gain:} 
MCR\textsuperscript{2} can be seen as a generalized form of {\em Information Gain} (IG). IG aims to maximize the reduction of entropy of a random variable, for example $\bm z$, w.r.t. an observed attribute, say $\bm \pi$:
$
\max_{\bm \pi} \; \mbox{IG}(\bm z, \bm \pi) \doteq H(\bm z) - H(\bm z \mid \bm \pi),
$
This is a measurement of the {\em mutual information} between $\z$ and $\bm \pi$ \cite{A1-17}.

This maximal information gain technique is widely applied with decision trees and other areas. However, MCR\textsuperscript{2} is used differently in several ways:
\begin{enumerate}
\item When the class labels $\bm \Pi$ are known, MCR\textsuperscript{2} focuses on learning representations $\bm z(\theta)$ rather than fitting labels.
\item In traditional settings of IG, the number of attributes in $\bm z$ cannot be so large and their values are discrete (typically binary). In this case, the “attributes” $\bm \Pi$ represent the probability of a multi-class partition for all samples. Their values can even be continuous.
\item As mentioned above, entropy $H(\bm z)$ and mutual information $I(\bm z, \bm \pi)$~\cite{hjelm2018learning} are not well-defined for degenerate continuous distributions. On the other hand, the rate distortion $R(\bm z, \epsilon)$ can be accurately and efficiently computed for (mixed)  subspaces. 
\end{enumerate}

\subsection{Properties of MCR\textsuperscript{2}} 

The MCR\textsuperscript{2} criterion~\eqref{eqn:maximal-rate-reduction} is theoretically very generalizable. It should be applicable to representations $\Z$ of any distribution with any attributes $\bm \Pi$, providing the rates $R$ and $R^c$ for the distributions can be evaluated accurately and efficiently. The optimal representation $\Z^*$ and partition $\bm \Pi^*$ have some interesting properties. \cite{A3} describe several useful properties of $\Z^*$ in the special case of subspaces -- a case very important to machine learning. When the desired representation for $\Z$ is multiple subspaces, the rates $R$ and $R^c$ in Eq.~\eqref{eqn:maximal-rate-reduction} are given by~\eqref{eqn:coding-length-eval} and~\eqref{eqn:compress-loss-eval}, respectively.

At the maximal rate reduction, MCR\textsuperscript{2} achieves optimal representation, denoted as $\Z^* = \Z_1^*\cup \cdots \cup \Z_k^* \subset \Re^d$ with $\rank{(\Z^*_j)}\le d_j$. $\Z^*$ has the following desired properties\footnote{See Appendix A (especially A.5) in~\cite{A3} for a formal statement and detailed proofs.}.

\begin{theorem}[Informal Statement~\cite{A3}]
Suppose $\Z^* = \Z_1^*\cup \cdots \cup \Z_k^*$ is the optimal solution that maximizes the rate reduction, as described in~\eqref{eqn:maximal-rate-reduction}. The following properties hold:

\begin{itemize}
\item \emph{Discriminative Between Classes}: 
If the ambient space is large enough ($d \ge \sum_{j=1}^{k} d_j$), the subspaces will all be orthogonal to each other, {\em i.e.} $(\Z_i^*)^{\top} \Z_{j}^* = \bm 0$ for $i \not= j$.
\item \emph{Maximally Diverse Representations}: 
As long as the coding precision is high enough ($\epsilon^4 < \min_{j}\left\{ \frac{m_j}{m}\frac{d^2}{d_j^2}\right\}$), each subspace achieves its maximal dimension, i.e. $\rank{(\Z^{*}_{j})}= d_j$. In addition, the largest $d_j - 1$ singular values of $\Z^{*}_{j}$ are equal. 
\label{thm:MCR2-properties}
\end{itemize}
\end{theorem}

Put another way, the MCR\textsuperscript{2} criterion encourages the data to become embedded into independent subspaces. The features of the data tend to be isotropic within the subspace. Moreover, the MCR\textsuperscript{2} criterion is eager to use any dimensional space available to it, with embeddings tending to use the \emph{highest} number dimensions as possible within the ambient space. This is marked difference from the IB objective~\eqref{eqn:information-bottleneck}.

\begin{figure}[!tbh]
	\centering
	\includegraphics[scale=0.27]{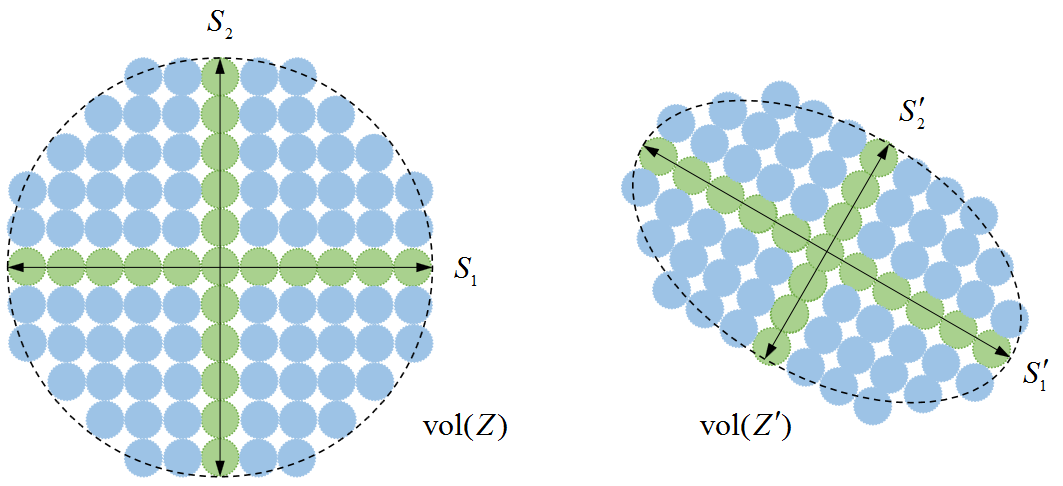}
	\caption{Two representations learned via reduced rates $\Z$ and $\Z'$. $R$ is the total number of $\epsilon$-balls packed into the joint distributions. $R^c$ is the number of $\epsilon$-balls (green) in the subspaces. $\Delta R$ is the difference - the blue balls. The MCR\textsuperscript{2} criterion~\cite{A3} encourages $\Z$.} 
	\label{fig:sphere-packing}
\end{figure}

\noindent \textbf{Comparison to the geometric OLE loss:} 
The {\em Orthogonal Low-rank Embedding} (OLE) loss was proposed by \cite{lezama2018ole} as a way of encouraging the de-correlation of the class features of different classes. Their idea was to try and maximize the maximum $\Delta R$, the difference between the nuclear norm of the whole distribution, $\Z$, and its subsets, $\Z_j$. Formally put, 
\begin{align}
	\max_{\theta}\,
\mbox{OLE}(\Z(\theta), \bm \Pi) \doteq  \|\Z(\theta)\|_* - \sum_{j=1}^k \|\Z_j(\theta)\|_*.
	\label{eqn:ole-loss}
\end{align}

\noindent They proposed that this loss could be a regularizer to be used alongside the cross-entropy loss, see Eq.~\eqref{eqn:cross-entropy}.

In Eq.~\eqref{eqn:ole-loss}, $\|\cdot \|_*$ refers to the nuclear norm, a non-smooth convex surrogate for low-rankness. Of course, the non-smoothness poses difficulties when trying to use gradient descent to learn features. $\log\det(\cdot)$, on the other hand, is {\em smoothly concave}.

OLE is always negative, achieving a maximum value of $0$ in the case where all subspaces are orthogonal. It pays no attention to dimensions. Comparing with $\Delta R$, the OLE loss serves as a geometric heuristic and does not promote diverse representations. Quite the opposite, in fact. OLE usually promotes the learning of {\em single-dimensional} representations for each class. This is in stark contrast to MCR\textsuperscript{2}, which encourages the learning of subspaces with maximal dimensions.

\noindent \textbf{Relation to contrastive learning:}
Contrastive Learning is a technique aimed at building good feature representations by ensuring that representations of samples from different classes are very different from each other, while representations of samples from the same classes are very similar \cite{hadsell2006dimensionality,oord2018representation,he2019momentum}. Moreover, this technique can be used even in the absence of class labels. Consider drawing a randomly chosen pair of samples $(\x_i, \x_j)$ from a dataset $\X$ with $K$ classes. The probability that the samples are from the same class is $1/K$. Thus, as the number of classes increases, the chance that two randomly drawn samples are from separate classes approaches certainty. 

However, for contrastive learning to work, a useful metric of “distance” between learned representations must be found. Using rate reduction one could say that, given a pair of sample representations $\Z_i$ and $\Z_j$, $\Delta R_{ij} = R(\Z_i\cup \Z_j, \epsilon) - \frac{1}{2}(R(\Z_i, \epsilon) + R(\Z_j, \epsilon))$ is just such a distance metric. It is then just a matter of increasing the distance between pairs (since they likely belong to different classes). 
Theorem~\ref{thm:MCR2-properties} states that the averaged rate reduction $\Delta R_{ij}$ is maximized when features from different samples are uncorrelated $\Z_i^\top \Z_j = \bm 0$ (see Figure~\ref{fig:sphere-packing}). Moreover, when comparing two different representations of the same underlying sample\footnote{This can be done by using multiple random augmentations.}, the features should be highly correlated, thus reducing the distance between them. 

Thus, MCR\textsuperscript{2} proves itself to a be a very natural fit for contrastive learning. Moreover, MCR\textsuperscript{2} can be used with more than a pair of samples. In fact, it can be scaled to increase or decrease the distance between an arbitrary number of representations, as long as they are likely from different (or the same) class.

\subsection{Summary}
\label{sec:MCR2Summary}

\begin{itemize}
\item This section introduced {\em Maximal Coding Rate Reduction} (MCR\textsuperscript{2}), an information-theoretic measure that can maximize the coding rage difference between an entire dataset and the sum of its individual classes.
\item This section described MCR\textsuperscript{2}'s relationship with a host of similar concepts, including Contrastive Learning, Information Gain, Cross-Entropy and OLE. It further discussed the theoretical guarantees (Theorem~\ref{thm:MCR2-properties}) that it will learn diverse and discriminative features.
\item This section showed how the coding rate can be calculated from practical data consisting of finite datasets with potentially degenerate subspace-like distributions. It can learn intrinsic representations in a host of settings including supervised, semi-supervised, unsupervised and self-supervised.
\end{itemize}

\section{Conclusion Remarks}
\label{sec:ConclusionRemarks}

\subsection{Segmentation via Minimum Lossy Coding Length}

Section~\ref{SegmentationClusteringIntro} introduced the concept of segmenting or clustering multi-subspace data using the rate-distortion principle and the measurement of \emph{minimum lossy coding length}. This technique sought to find the shortest encoding length for data, subject to a certain acceptable level of distortion. This was to allow for the optimal segmentation of mixed Gaussian and subspace-like data.

Other estimates, including LMDL, LML, MDL and ML were also discussed in comparison to the \emph{minimum lossy coding length}. These were shown to be optimal only in the theoretical case where there was an infinite number of i.i.d. samples from a given class distribution -- the {\em asymptotic} case.

\emph{Minimum lossy coding length} was shown to be more suitable to real-world situations, with finite (and often quite small) numbers of data samples available, approximating an (often degenerate) Gaussian source. The technique was shown to be able to give a \emph{tight upper bound} in both deterministic and probabilistic segmentation settings\footnote{\cite{A1} proved that deterministic segmentation is approximately asymptotically optimal. See Section 4.2 of their paper.}.

In Section~\ref{sec:MCL}, this technique was further developed with the introduction of an efficient data-driven bottom-up algorithm (Algorithm~\ref{Algo-1}), which was shown to have the following features:

\begin{itemize}
\item A physically meaningful quantity, e.g. binary bits, can be measured to evaluate the gain or loss of segmentation. Moreover, an explicit formula exists for evaluating the coding rate or coding length, which allows one to judge the suitability of a segmentation. Given that ability, it then possible to investigate the effect of varying the distortion.
\item The algorithm requires no initialization or previous knowledge of the underlying Gaussian models or subspaces. Instead, it is a {\em greedy} algorithm that operates by iteratively merging small subsets into larger ones, each time attempting to improve its score.
\item The algorithm exhibits {\em polynomial} scaling in both number of samples and data dimension. So long as the distortion is reasonable in comparison to the sample density, it is likely to converge on the optimal solution. And it is very robust to both distortions, noise and outliers.
\end{itemize}

This technique has great potential for further extension into other ares, such as detection, classification and recognition. Moreover, its segmentation ability may be further applicable with other types of structures, such as {\em non-Gaussian} probabilistic distributions and {\em non-linear manifolds}. Extending the efficiency and speed of the greedy algorithm is expected to be a fruitful direction of future research.

\subsection{Classification via Minimum Incremental Coding Length}

Section~\ref{sec:ClassificationICLIntro} introduced the concept of the \emph{Minimum Incremental Coding Length} (MICL), described in Criterion~\ref{Criterion-1}. This attempts to assign new test samples to whichever class can be shown to require the smallest increase in bit count to code the sample, subject to an allowable distortion level.

Section~\ref{sec:AnalysisofMICL} showed the {\em asymptotic optimality} of the MICL criterion with Gaussian distributions. MICL was also compared to MAP (\ref{sec:map}) and SVM (Section~\ref{sec:KernelMICL}) and the relationships analysed. MICL was shown to extend the working conditions of these other more classic approaches and apply them to situations where the distribution is {\em singular} in a high-dimensional space or the sample set is {\em sparse}. When applied to a classifier (Algorithm~\ref{Algo-2}), there were several positive traits:
\begin{itemize}
\item MICL shows a {\em regularization} effect on the density estimate when dealing with a small number of samples. Theorem~\ref{them:AsymptoticMICL} showed that MICL is equivalent to the regularization version of MAP or ML in the asymptotic case. However, MICL has a {\em rewards effect} on relatively higher dimensional distributions. Theorem~\ref{them:ConvergRateofMICL} showed that MICL's convergence becomes even better when the distribution is closer to singular.
\item MICL can be applied in practice with very little pre-processing or engineering of the data to be classified. This is because MICL has been shown to be able to automatically exploit low-dimensional data structures lurking within high-dimensional data.
\end{itemize}

\subsection{Representation via Maximal Coding Rate Reduction}

Section \ref{sec:How2learnDDR} introduced the the criterion of \emph{Maximal Coding Rate Reduction} (MCR\textsuperscript{2}). This is an information-theoretic measure that sought to maximize coding rate difference between encoding an entire dataset versus the sum of encoding each class-specific subset of the dataset. This was shown to be able to identify the \emph{intrinsic low-dimensional structures} existing within \emph{high-dimensional data} that can best be used to discriminate between different classes, this being very useful as part of clustering and classification of data. The following points were covered:
\begin{itemize}
\item Existing frameworks, including Constrastive Learning, Cross-Entropy, Information Gain and OLE loss, were examined and their relationship with MCR\textsuperscript{2} explored. Theorem~\ref{thm:MCR2-properties} was introduced, which provided theoretical guarantees about the learning of {\em diverse} and {\em discriminative} features.
\item MCR\textsuperscript{2} was shown to useful in a large number of settings, including supervised, semi-supervised, unsupervised and self-supervised contexts.
\item MCR\textsuperscript{2} was shown to be able to accurately compute the coding rate with finite samples from degenerate subspace-like distributions. The criterion was demonstrated with mixed subspaces and shown to be extendable to {\em any arbitrary} mixed distributions or structures.
\item MCR\textsuperscript{2} provides good explanations for the utility of many existing frameworks and heuristics used throughout the deep learning literature.
\item There is still a lot of room for future research, including questions such as: {\em (i)} Why is it robust to label noises in the supervised setting? {\em (ii)} Why are features self-learned with MCR\textsuperscript{2} alone are effective for clustering? Much work has been done recently \cite{Zhang2021,Li2022,Tong2022}, including attempts to make MCR\textsuperscript{2} more computationally efficient \cite{vmcr2}.
\end{itemize}

MCR\textsuperscript{2} has great potential to provide a {\em principled} and {\em practical} objective for many deep learning tasks. It may lead to better deep learning architectures and new understanding of the theory behind training tasks. It even provides the chance to analyse a deep learning stack as a “white-box”, for example by monitoring the amount of rate reduction $\Delta R$ gained in each layer of a deep network \cite{A5redunet}.

\section*{Acknowledgements}

The authors would like to thank Professor Yi MA from Department of EECS, University of California, Berkeley and his team for their kindly inspiration and guidance.
And we would also like to thank the anonymous reviewers for their comments and suggestions.
\\

{\noindent{\textbf{Contributors~{\tiny{(refer to https://www.casrai.org/credit.html)}}}}}

{\footnotesize Kai-liang LU designed the summary note and drafted the manuscript. Avraham Chapman helped organize the manuscript. Avraham Chapman and Kai-liang LU revised and finalized the paper.
}

{\noindent{\textbf{Compliance with ethics guidelines}}}

{\footnotesize Kai-liang LU and Avraham Chapman declare that they have no conflict of interest.
}

\bibliographystyle{fitee}
\bibliography{reference}

\end{document}

%% file: TexFiles/math_notation.tex
\newcommand{\trace}{\mathrm{tr}}

\newcommand{\Appendix}[1]{the full version for}

\newtheorem{criterion}{Criterion} %

\newcommand{\x}{\bm{x}} %
\newcommand{\y}{\bm{y}} %
\newcommand{\z}{\bm{z}}

\newcommand{\I}{\bm{I}}

\newcommand{\N}{\mathcal{N}} %

\newcommand{\R}{\mathbb{R}}  %
\renewcommand{\Re}{\mathbb{R}}

\newcommand{\X}{\bm{X}} %
\newcommand{\Z}{\bm{Z}} %
\newcommand{\rank}{\textup{\textsf{rank}}}

\newcommand{\tr}{\textup{\textsf{tr}}}

\newcommand{\cL}{\mathcal{L}}

\newcommand{\cT}{\mathcal{T}}

\DeclareMathOperator*{\argmin}{arg\,min}
\usepackage{pifont}